\documentclass[letterpaper, 10 pt, conference]{ieeeconf}  

\IEEEoverridecommandlockouts                              
\usepackage{graphicx} 
\usepackage{tablefootnote}

\title{\LARGE \bf
Understanding Career Progression in Baseball Through Machine Learning*
}

\author{Brian Bierig,$^{1}$ Jonathan Hollenbeck,$^{2}$ and Alexander Stroud$^{3}$
\thanks{*This work was not supported by any organization}
\thanks{$^{1}$Department of Management Science and Engineering, Stanford University
    {\tt\small bbierig@stanford.edu}}%
\thanks{$^{2}$Professional Student in Computer Science, Stanford University,
	{\tt\small jonoh@stanford.edu}}%
\thanks{$^{3}$Undergraduate in Mathematical and Computational Sciences, Stanford University,
	{\tt\small astroud@stanford.edu}}%
}

\begin{document}

\maketitle
\thispagestyle{empty}
\pagestyle{empty}

\begin{abstract}

Professional baseball players are increasingly guaranteed expensive long-term contracts, with over 70 deals signed in excess of \$90 million, mostly in the last decade.  These are substantial sums compared to a typical franchise valuation of \$1-2 billion.  Hence, the players to whom a team chooses to give such a contract can have an enormous impact  on both competitiveness and profit.  Despite this, most published approaches examining career progression in baseball are fairly simplistic.  We applied four machine learning algorithms to the problem and soundly improved upon existing approaches, particularly for batting data.
\end{abstract}

\section{INTRODUCTION}

The typical mode of entry for a player into baseball is through the first-year player draft.  Players usually enter the draft immediately after high school or college and then spend several years in the drafting team's minor league system.  When deemed ready, the drafting team can promote the player to the Major Leagues.  From that time forward, the team retains rights to the player for 6 years, with salaries determined by an independent arbiter.  After six years, the player can elect to become a free agent and is free to negotiate a contract with any organization.  It is typically at this point when the largest contracts are given out.

We looked at all players who appeared in at least 7 seasons since 1970.  We used all the data from these players' first 6 seasons as features, both separated by season and aggregated together.  We combined this data with additional exogenous information, such as age of first Major League season, handedness, and fielding position and then attempt to use machine learning to predict a player's career trajectory, as represented by yearly wins-above-replacement, the gold-standard statistic for player value.  We then compared four machine learning models: linear regression, artificial neural networks, random forest ensemble, and $\epsilon$-support vector.  Each model was trained twice, once for batters and once for pitchers.

\section{RELATED WORK}

The available literature on baseball career progression provides scant help for examining the problem with machine learning.  A simple, yet oft-cited, approach is the ``delta method,'' which appears to have originated from baseball statistician Tom Tango\cite{c2}.  To estimate change in some metric between two ages, the metric is calculated for all players active in a chosen time period who played at those two ages.  The values at the two ages are then summed and the average difference is taken.  As a specific example, hitters who played in the major leagues at both age 30 and age 31 saw their WAR decrease on average by 0.28 from their age-30 season to their age-31 season. Thus, if a player had a WAR value of 2 in their age-30 season, the Delta Method predicts that they will have a WAR of 1.72 in their age-31 season. If their WAR value was 1 in their age-30 season, their predicted age-31 WAR is 0.72. When deltas are computed across all ages, a complete ``aging curve'' can be drawn. The Delta Method depends only on a player's age and WAR in their most recent season. Given the relatively simple structure of this model, we believed a more complex machine learning approach could improve upon its predictions.

Others have similarly elaborated on the basic Delta Method model.  Fair (2008)\cite{c2} used a nonlinear fixed-effects regression model to generate aging curves, where the periods of upslope and of decline were separate but constrained to merge at the peak age.  Bradbury (2009)\cite{c3} used a multiple regression model, where average career performance was included as a variable.  Lichtman (2009)\cite{c4} used the delta method but attempted to add corrections for the ``survivor bias,'' the phenomenon occurring when a player's performance degrades so much in a year that he does not play much the next.  The Bradbury piece is the outlier in the literature, finding that players peak at age 29, two years later than most other analyses.  That being said, all studies had different thresholds for including a player in the analysis, included data from different time periods, and utilized different approaches for normalizing between years, leagues, and ballparks.  These factors make comparisons difficult. 

We note that the above discussion only includes publicly published studies on the topic of career progression.  It is possible and likely that some of the 30 professional baseball teams have performed proprietary research on the topic.

\section{DATASET AND FEATURES}

We obtained our dataset from Kaggle\cite{c6} and scraped WAR values from Baseball Reference\cite{c9}.  The Kaggle format aggregates all data for each ``stint'', which consists of a player's contiguous activity with a team over a unique year.  For example, if a batter played for team A for the first half of 2015, then team B for the second half of 2015 and all of 2016, they would have three stints -- one for team A, and two for team B.

The goal of our model is to allow teams to evaluate potential contracts with players who are free agents for the first time.  With this in mind, we treated all the information we have about each player for the first six years as source data, and everything from the years following as outcomes or y values.   To set a baseline, we aggregated all the data from years 1-6 and calculated simple metrics, then trained against various y values from the following years. In later models, we included separate features from each year in addition to the aggregates.  We also decided to use WAR as the sole evaluation metric.  A player was thus represented by a feature vector with yearly and aggregate information, which we used to predict their WAR in following years.

Some players lacked sufficient activity to justify a prediction, or did not receive a contract after the end of team control.  To adjust for this, we excluded all players with careers spanning under seven years. For batting predictions, we also filtered out years where the player had 0 at bats, since our dataset includes a batting stint for all players with any pitching or fielding activity.  We then excluded batters who did not have at least 1 active batting year both before and after the 6 year mark.  Similarly, we also filtered out pitchers who didn't appear in at least 1 game both before and after the 6 year mark.

Batting predictions for pitchers are potentially important, but difficult to predict because of quirks in the league structure.  Additionally, many batters had intermittent pitching data that should not factor into any analysis.  To address these issues, we excluded all batters with enough data to show up in our pitching analysis.  This was fairly aggressive; future work could include a more careful approach.
																
Finally, we excluded all players with a start year before 1970.  The league has changed over the years, and we found that earlier data struggled to help predict contemporary outcomes.  Even though we eliminated a large proportion of contemporary players, we still included a large percentage of the overall data (see tables 1 and 2 below).

We utilized one-hot-encoding to structure features for categorical data such as decade, position, and handedness.

Using WAR as the prediction metric raised two issues.  First, there are competing definitions and interpretations of WAR, each of which have their own dataset.  We discuss this later.  Second, once players drop out of the league, they no longer record a WAR value.  For these ``missing'' years, we set WAR to 0 since they're presumably providing the same value as at-replacement-level players (who are by definition interchangeable with the best minor league players).  This is not necessarily indicative of skill, since the player would likely produce a negative WAR if their career ended for performance reasons.  Furthermore, consistent players with small, positive WAR could still be worth substantial medium term contracts, so our model represents the value of marginal players poorly.  We tried setting missing WAR to arbitrary small negative values (like -0.5 and -1) and observed slight improvements to all models (including the baseline delta method), but felt that this was not worth the additional model complexity.

Because WAR measures player value in a contemporary context, we attempted to normalize our other features in a similar fashion.  For example, a .270 batting average would be .02 above the mean in 1970, but average in 2000.  However, applying this on a year-to-year basis substantially decreased model performance.  We believe that this was due to large noise in mean variations, particularly for features that reflect rare events (like home runs).  Alternatives continued to hurt model performance (locally weighted means) or had marginal effects (per-decade weighting).

Before use in models, we applied min-max mean normalization to all features with scikit- learn's MinMaxScaler package.\cite{c5}

\begin{table}[!htbp]
\begin{center}
\caption{Cleaning Batter Data}
\begin{tabular}{|c|c|c|c|c|} \hline
& Contemporary & Included & Percent Included \\ \hline
Unique Players & 7956 & 1669 & 21.2 \\ \hline
Total ABs (K) & 6050 & 5167 & 85.4 \\ \hline
\end{tabular}
\end{center}
\end{table}

\begin{table}[!htbp]
\begin{center}
\caption{Cleaning Pitcher Data}
\begin{tabular}{|c|c|c|c|} \hline
& Contemporary & Included & Percent Included \\ \hline
Unique Players & 4395 & 1390 & 31.6 \\ \hline
Total IPOUTs (K) & 4773 & 3831 & 80.3 \\ \hline
\end{tabular}
\end{center}
\end{table}

\section{METHODS}

As noted, each training example contained over two hundred features, many of which were highly correlated.  For example, the only difference between ``at-bats'' and ``plate appearances'' in a given year is that the former includes walks and a few other rare occurrences.

To account for these effects, scikit-learn's Recursive Feature Elimination (RFE) tool[5] was utilized, along with a basic linear ridge model with regularization ``alpha'' parameter set to 2.  This utility instantiates the model with all features and then eliminates, upon each iteration, a user-defined number of features with the lowest coefficients.  With our relatively small dataset, we chose to only eliminate one feature per iteration.

The utility was run for each year we attempted to predict (7-11) and for both batters and pitchers.  Moreover, for each one of these items, the RFE was run several times, so as to gauge the performance of the model with various different numbers of features.  Near-optimal model performance could be achieved with just a few features; however, the results were fully asymptotic after around 15-20 features.  For all future tests, 15-20 features were utilized.

After selecting the optimal features, we ran the data through four different models as described below.

\subsection{Linear Regression with L2 Regularization (Ridge Model)}

This is a generalized linear model of the normal distribution.  A weight for each parameter is determined by maximum-likelihood estimation over the training set.  The equation of the cost function is given below.

\[
J(\theta) = \frac{1}{2m} \bigg[ \sum_{i=1}^m \Big( h_\theta(x^{(i)}) - y^{(i)} \Big)^2 + \lambda \sum_{j=1}^n \theta_j^2\bigg]
\]

Because this model is prone to overfitting, a regularization term was also added to the objective.  This puts downward pressure on the size of each feature's coefficient, balanced against finding the optimal value to fit the training set.

\subsection{Multi-Layer Perception Regression (Neural Network)}

A fully connected artificial neural network consists of layers of neurons, where every neuron is connected by a weight to all of the neurons in the previous layer.  As discussed later, models with 1 or 2 layers were examined.  At each hidden neuron, the rectified linear unit (ReLU) activation function was used.  Since the model was utilized for regression, the output node was not transformed.  Due to the relatively small size of our dataset, the Newtonian 'L-BFGS' solver was used.  The loss function for this model is the sum of the squared error for each training example, similar to that defined for Linear Regression.

\subsection{Random Forest Regression (Tree Bagging Model)}
The random forest model\cite{c7} utilized 100 separate decision trees.  For each decision tree, samples were selected by bootstrapping the training set.  Each decision tree was then formulated by splitting the data along the features that explain the most variation.  Results were then averaged to produce the final model.  Note that the model was also tested by restricting the number of features considered at each tree split; however, this generally produced worse results, so the final model did not contain the restriction.  By convention, then, the final model was a bagging model and not a random forest.

\subsection{$\epsilon$-Support Vector Regression (SVR)}

$\epsilon$-SVM, or SVR, is an extension of the support vector machine concept to regression problems.\cite{c8}  The algorithm attempts to find the response function that best fits the training data to the y-values without deviating by more than $\epsilon$.  A regularization-like term called the Box coefficient is also included in the optimization in order to handle training examples that exist outside the $\epsilon$ region, similar to the purpose of the hinge loss equation in classification problems.  The Box coefficient also reduces overfitting.  The cost function is provided below, where $\xi$ and $\xi^*$ are slack variables to the $\epsilon$ constraints.  In our experiments, the SVR model was used exclusively with a Gaussian kernel.

\[
J(\beta) = \frac{1}{2}||\beta|| + C \sum_{i=1}^{m} (\xi_i + \xi_i^*)
\]

\subsection{Hyperparameter Optimization}

For each of the four models above, scikit-learn's grid search utility was used to optimize hyperparameters.  Recall that we included data for two types of players (batters and pitchers), five prediction years, and four models.  As a result, grid-search was utilized 40 times, even though results tended to remain somewhat similar on each iteration.  The parameters for each model and the range over which they were optimized are listed in the table below.

\begin{table}[!htbp]
\small
\begin{center}
\caption{Parameter Ranges}
\begin{tabular}{|c|c|} \hline
Model & Hyperparameters \\ \hline
Neural Net & \begin{math}\alpha\end{math}\begin{math}\in\end{math}[0.01, 100], lay-1\begin{math}\in \end{math}[4, 16], lay-2\begin{math}\in \end{math}[0, 5] \\ \hline
Random Forest & max depth \begin{math}\in \end{math} [2, 7], min leaf split \begin{math}\in \end{math} [1, 4] \\ \hline
\begin{math}
\epsilon-SVM (rbf)\end{math} & \begin{math}
\epsilon\end{math}\begin{math}\in \end{math}[1e-4, 1e2], C\begin{math}\in \end{math}[0, 10e5], \begin{math}\gamma\end{math}\begin{math}\in \end{math}[1e-5, 1e2]\\ \hline
Ridge\tablefootnote{Note that sci-kit learn uses $\alpha$ instead of $\lambda$ to refer to the regularization coefficient.} & \begin{math}\alpha\end{math} \begin{math}\in \end{math} [0.01, 100] \\ \hline
\end{tabular}
\end{center}
\end{table}

All feature selection was conducted on a training set consisting of 80\% of player data.  After features were set, model hyperparameters were chosen using the training set and 3-fold cross validation.  This approach minimized overfitting, while still allowing full use of the training set. Each trained model was then used to evaluate the unseen 20\% test set.

\section{RESULTS AND DISCUSSION}

The results on the test set, as measured by explained sum of squares (R$^2$), are displayed in the figure below.  There was no clear winner between the machine learning models.  For batters, the models were all typically able to explain approximately 60\% of variance.  This varied moderately based on year.  For example, year 9 -- the third year we attempted to predict -- had an R$^2$ closer to 0.55 for all models.  We suspect this is indicative of random fluctuation.

\includegraphics[width=\linewidth]{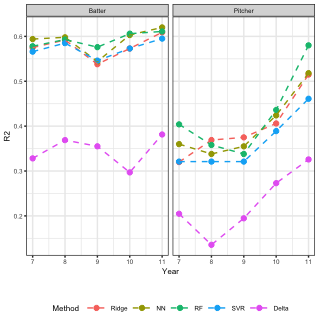}

For pitchers, the models produced results explaining between 30\% and 40\% in years 7-10, substantially lower than for batters.  This was most likely caused by another form of variation for pitchers -- injuries.  Indeed, many pitchers at some point in their career will tear their ulnar collateral ligament (UCL).  This injury requires surgery, and the surgery requires at least a year of healing.  Even with this surgery, some pitchers never recover their previous level of performance.  Although injuries also occur to batters, there is no equivalent that is as frequent or devastating; the typical injuries for batters are also injuries that can happen to pitchers.

We also note that model performance actually improved in year 11 for pitchers.  The cause of this improvement is simple: by year 11, the models are projecting that certain pitchers will no longer be active.  When this assumption is valid, all the variation is explained, whereas in previous active years there would still be some deviation.

For batters in years 7 and 8, two features -- cumulative WAR in first 6 seasons and WAR in 6th season -- fared essentially as well in predicting future performance as any combination of additional statistics.  By year 9, age when entering the league was also needed to reach the maximal explained sum of squares.  In years 10 and 11, age was more critical than sixth year WAR for predicting performance.  This presumably has two causes: (1) players entering the league at more advanced ages will fare worse in their 10th season and (2) performance in year 6 says more about performance in year 7 than it does about year 10.  The idea that a couple performance metrics are the best predictors of late career longevity lends some credence to Bradbury's approach, described in Related Work.  Bradbury utilized a ``delta'' approach but controlled for baseline performance level.

Feature optimization for pitchers was similar to batters.  In general, cumulative WAR in the first 6 seasons and individual WAR in the 6th season were sufficient for building good models.  In year 8, strikeouts in the 6th season actually beat out 6th season WAR for second place in the recursive feature elimination; however, this was simply another indicator of value in the 6th year.  More interestingly, age never showed up as critically in the model as it did for batters.  Rather, additional explanatory power in later years was gained through statistics such as cumulative complete games in the first 6 seasons and some parameters of the first season (total batters faced, walks).  The former, although seemingly an unimportant statistic, has some justification.  Truly elite players, who tend to have long careers, also tend to throw more complete games.  The complete game stat may also have been differentiating between starting pitchers and relief pitchers, who were both included in the pitching analysis.  Relief pitchers, by role definition, will never be able to throw a complete game.  The importance of first year performance is curious and warrants more study.

Interestingly, some features we thought would weigh heavily had minor impact.  It had been our expectation that the decade in which a player's career began might be crucial, especially given that our dataset included the steroid era of the late 1990s.  None of the one-hot encoded data corresponding to decade, however, appeared as a top feature.  Similarly, we hypothesized that positional data might be crucial.  Specifically, ``up-the-middle'' players (shortstop, second base, and centerfield) are widely regarded as the most athletic and might age better.  Although the third base and left-field positions occasionally appeared as negative weights in the linear model, these were never more than the 30th most important feature.  Similarly, we found that height and weight were never critical variables.  Part of this could be due to the structure of our dataset; all players were assigned a single value for their entire careers, which is unrealistic for weight.  Handedness (right, left, and switch) was also not deemed important.  However, we had no reason to expect that it would be. 

Even though only a few features were sufficient for reaching maximal explanatory power in our models, they still handily beat the naive ``delta method'' described earlier.  To be fair, however, most authors who utilize the delta method (described in Related Work) believe it to be insufficient on its own and will perform corrections for aspects such as survivor bias, injuries, time period, and baseline talent level.  We performed no such adjustments, with the intention to let the machine learning models infer these relationships.  As such, the argument could be raised that our delta method calculation set up a proverbial ``straw man'' data point.  Further analysis should attempt to directly compare our models with those of Bradbury, Fair, and Lichtman; however, this is not immediately possible due to different player inclusion criteria.

A comparison of the performance of the neural network to the delta method is provided in the plot below.  The plot compares the actual results of the players in the test set with the values predicted by the models.  Color is used to display frequency of observation -- hence, the lighter blue color represents the majority of observations which are heaviest around zero and extend upwards along the red reflection line.  The darker blue points have low frequency of observation and tend to extend further away from the reflection line.  Clearly, the neural network model has a spatially tighter distribution of values for both hitters and pitchers compared to the delta method.  These results are representative of the other machine learning models.  We also observe that the delta method has more predictions in negative territory.  This is unrealistic, in the sense that these players would likely be benched; however, the naive assumptions in the model do not account for this.  Note that all five years of data and predictions were included to generate these heat plots.  The keen observer might also remark on the five points in the upper right of the neural network batter plot.  We note, after the fact, that baseball's all-time home-run leader (and a player who experienced his best seasons in his late 30's), Barry Bonds, was in our test set, and these points correspond to him.  It is interesting that the more complicated neural network model was able to maintain high predictions for Bonds for all 5 years in a way that the delta method was not.

\newpage
\includegraphics[width=\linewidth]{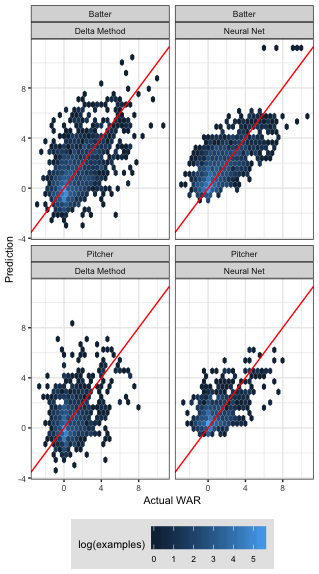}

\section{CONCLUSIONS AND FUTURE WORK}

In summary, we trained four machine learning models with data from the first 6 seasons of players careers, dating back to 1970.  We then attempted to predict a player's value, measured by WAR, in years 7-11.  For batters, we were typically able to forecast around 60\% of variation, while only 30\% - 40\% was possible for pitchers.  This is perhaps the largest takeaway from the study: if deciding between giving a long-term contract to a batter or a pitcher, all else being equal, the team should prefer the batter.  For both batters and pitchers, results handily beat the simplistic delta method calculation described in literature, although more work should be done to directly compare our results with the different variations other authors have attempted.  Lastly, we found that cumulative WAR over the first 6 seasons and 6th season WAR are the best predictors of later career performance.  For batters, rookie season age becomes critical starting in year 8.

Much time in this project was spent on feature selection; moving forward, more time should be spent on player inclusion criteria.  While baseball in the early 20th century was quite different from baseball today, the year 1970 was somewhat arbitrarily chosen as a dividing line to ensure that players from baseball's early eras would not contaminate our predictions.  A larger dataset (dating back to WWII or earlier) could be considered.  Moreover, we chose relatively lenient criteria for player inclusion (at least 7 years in league) so as to maximize total league at-bats included in the study.  The most popular literature criteria is 10 years, so this should also be attempted.  Lastly, there is some debate over the optimal calculation for wins-above-replacement.  In fact, three websites -- Baseball Reference, Fangraphs, and Baseball Prospectus -- tabulate similar, but competing, versions of the metric.  We utilized the Baseball Reference version due to ease of access; however, it would be worthwhile to compare the Fangraphs and Baseball Prospectus versions to see if results differ.

\addtolength{\textheight}{-12cm}   

\section{Contributions:}

Jonathan Hollenbeck led the data cleaning portion of the activities, setting up a SQL database system for easier manipulation.  Brian Bierig led the machine learning aspect of the project in sci-kit learn, with assistance and further analysis from Zander and Jonathan.  Zander Stroud also scraped all WAR data used in this project and implemented the delta method. Jonathan Hollenbeck created visualizations of model results. Brian handled most of the background research on related work. All contributors participated in discussions and brainstorming throughout the process.  

\end{document}